\documentclass{bmvc2k}

\usepackage{capt-of}   
\usepackage{graphicx}
\usepackage{wrapfig}   
\usepackage{amsmath,amssymb}
\usepackage{booktabs,multirow,adjustbox}
\usepackage[
linesnumbered,ruled,vlined]{algorithm2e}
\usepackage{cleveref}       

\usepackage{pifont}
\usepackage{verbatim,comment}
\usepackage{kotex}          
\newcommand{\cmark}{\ding{51}}
\newcommand{\xmark}{\ding{55}}

\usepackage{calc}   
\usepackage{makecell}
\title{Stabilizing Open-Set Test-Time Adaptation via Primary-Auxiliary Filtering and Knowledge-Integrated Prediction}

\addauthor{Byung-Joon Lee}{lbj2001@skku.edu}{1}
\addauthor{Jin-Seop Lee}{wlstjq0602@skku.edu}{2}
\addauthor{Jee-Hyong Lee$^{\dag,}$}{john@skku.edu}{1,2}

\addinstitution{
Department of Computer Science and Engineering,\\
Sungkyunkwan University
}

\addinstitution{
Department of Artificial Intelligence,\\
Sungkyunkwan University
}

\runninghead{LEE ET AL.}{Stabilizing Open-Set TTA
via PAF and KIP}

\begin{document}

\maketitle
\AddToShipoutPictureBG*{%
  \AtTextLowerLeft{%
    \put(0,-10pt){\makebox[\textwidth][l]{\scriptsize\dag\,Corresponding author.}}%
  }%
}
\begin{abstract}
Deep neural networks demonstrate strong performance under aligned training-test distributions. However, real-world test data often exhibit domain shifts. Test-Time Adaptation (TTA) addresses this challenge by adapting the model to test data during inference. While most TTA studies assume that the training and test data share the same class set (closed-set TTA), real-world scenarios often involve open-set data (open-set TTA), which can degrade closed-set accuracy. A recent study showed that identifying open-set data during adaptation and maximizing its entropy is an effective solution. However, the previous method relies on the source model for filtering, resulting in suboptimal filtering accuracy on domain-shifted test data. In contrast, we found that the adapting model, which learns domain knowledge from noisy test streams, tends to be unstable and leads to error accumulation when used for filtering. To address this problem, we propose \textbf{Primary-Auxiliary Filtering} (\textbf{PAF}), which employs an auxiliary filter to validate data filtered by the primary filter. Furthermore, we propose \textbf{Knowledge-Integrated Prediction} (\textbf{KIP}), which calibrates the outputs of the adapting model, EMA model, and source model to integrate their complementary knowledge for OSTTA. We validate our approach across diverse closed-set and open-set datasets. Our method enhances both closed-set accuracy and open-set discrimination over existing methods. The code is available at \url{https://github.com/powerpowe/PAF-KIP-OSTTA}.
\end{abstract}
\section{Introduction}
\label{sec:intro}
Deep neural networks have achieved outstanding performance across a wide range of tasks. However, this strong performance relies on the assumption that the training and test data are drawn from the same distribution \cite{croce2021robustbench}. In real-world scenarios, test data may undergo various domain shifts, such as changes in weather, brightness, or camera noise. To address this challenge, test-time adaptation (TTA) \cite{liang2024survey, wang2021tent, wang2022continual} aims to adapt the model to the test domain on the fly, during inference.

Most TTA studies have focused on the case where the classes in the training data and test data are identical (i.e., closed-set TTA) \cite{chen2022adacon, dobler2023rmt, brahma2023probabilistic, zhao2023delta}. However, it is common for unseen class data (i.e., open-set) to appear at test time \cite{geng2020opensetsurvey}. In such scenarios, the TTA model continually learns incorrect information about closed-set classes, leading to error accumulation and degraded accuracy \cite{gao2024unient}. Furthermore, both open-set and closed-set data may appear with domain shifts, making it challenging to distinguish between them.

To address this problem, recent studies have proposed Open-set Test-Time Adaptation (OSTTA) methods that adapt to test data containing both closed-set and open-set samples \cite{gao2024unient, lee2023ostta, gong2024sotta, yu2025stamp}. These methods estimate the confidence of each test sample, classify it as either open-set or closed-set, and leverage this categorization during adaptation.

\begin{figure}[t]
  \centering
  \subfigure[Percentage of wrongly filtered closed-set samples with UniEnt.\label{fig:unientlog}]{
    \includegraphics[width=.60\linewidth]{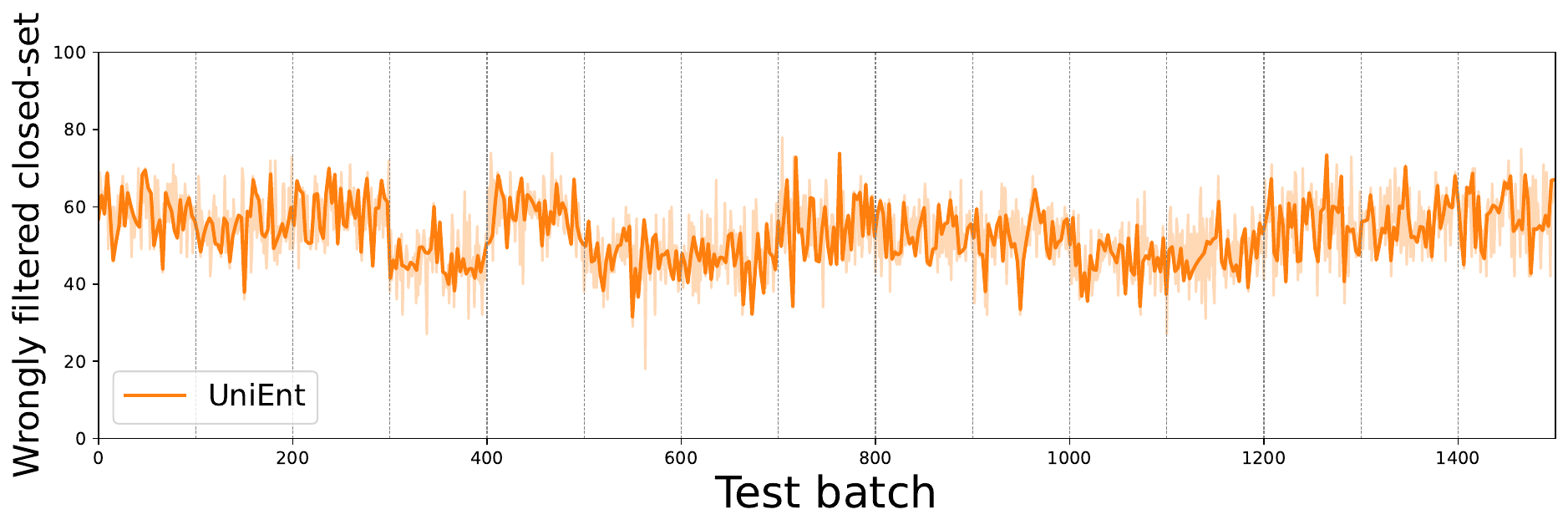}}
  \vspace{-1em} 
    
  \subfigure[Percentage of wrongly filtered closed-set samples with adapting, EMA, and our filter.\label{fig:ourslog}]{
    \includegraphics[width=.45\linewidth]{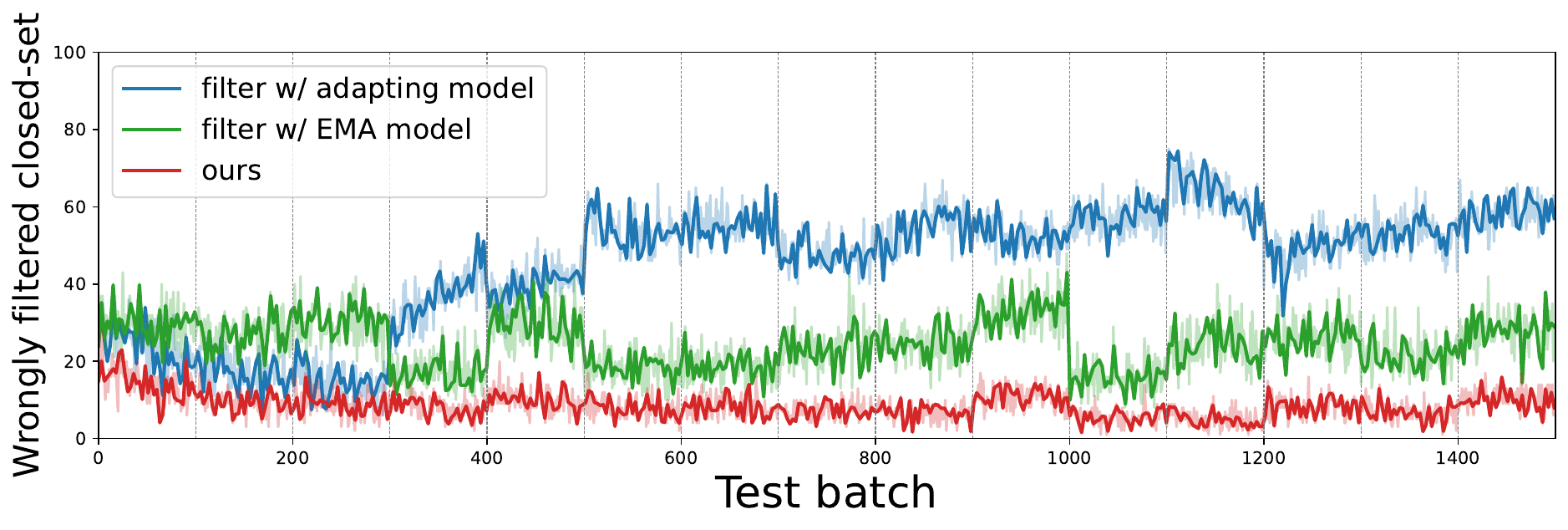}}
  \hfill
  \subfigure[H-score with adapting, EMA, and our filter.\label{fig:hs}]{
    \includegraphics[width=.45\linewidth]{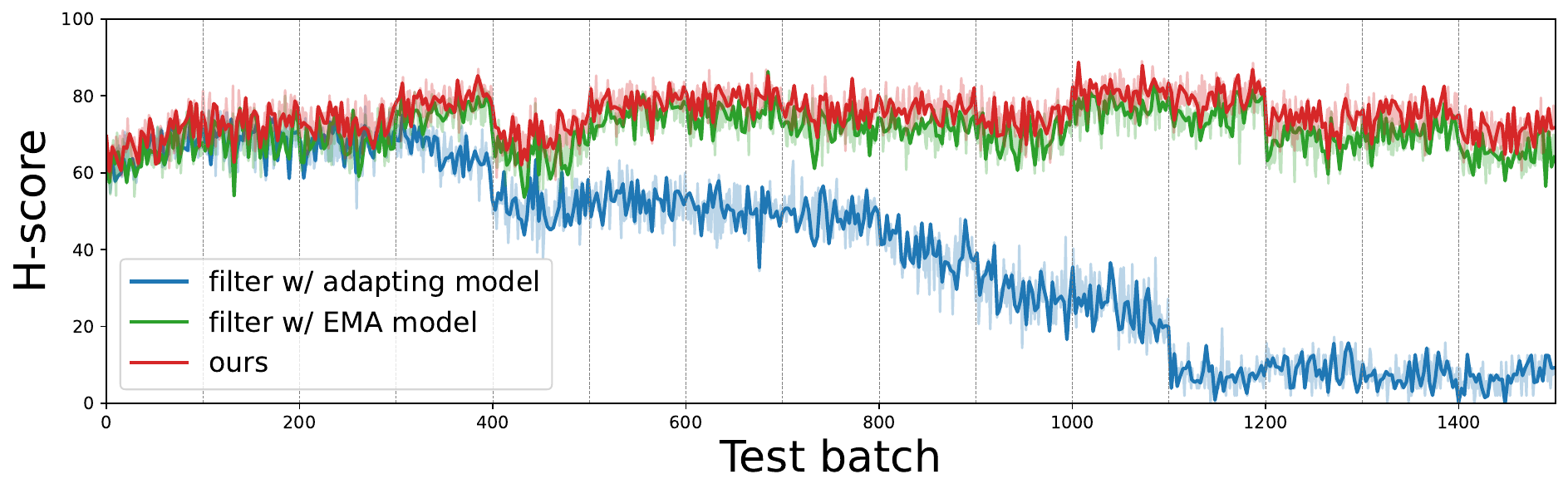}}

  \caption{Wrongly filtered closed-set samples and H-score on CIFAR100-C with Textures-C. The vertical dashed line marks domain shifts.}
  \vspace{-1.0em}
  \label{fig:motivation}
\end{figure}

While most OSTTA studies discard low-confidence data during adaptation, UniEnt \cite{gao2024unient} demonstrates that maximizing the entropy of such data can be effective. By increasing the entropy of open-set samples, the adapting model improves its ability to distinguish unknown instances by effectively learning what it does not know. Specifically, UniEnt utilizes a source pre-trained model to compute the similarity between each test sample and the source-domain class prototypes. Samples that closely align with known source classes are treated as reliable and optimized via entropy minimization, while those that appear dissimilar are considered unreliable and optimized via entropy maximization. 

Incorporating domain knowledge into filtering is crucial, as domain-shifted test samples often exhibit low confidence when evaluated by a source model not trained on the target domain. However, UniEnt overlooks domain knowledge in filtering since its confidence scores rely solely on features from the fixed source model and class prototypes. As a result, it fails to exploit test-domain information and incorrectly filters many closed-set samples as open-set, as shown in \cref{fig:unientlog}.

One way to leverage domain knowledge into filtering is to utilize the adapting model itself. However, relying on the adapting model's confidence for loss selection leads to significant error accumulation. This results in confirmation bias, where the adapting model reinforces its initial uncertainty by further reducing the confidence of low-confidence samples, as shown in \cref{fig:ourslog} and \cref{fig:hs}.

We found that the exponential moving average (EMA) model of the adapting model helps prevent error accumulation, but fails to promptly capture domain shifts at the current time step, thereby resulting in suboptimal filtering performance. To address this, we propose \textbf{Primary-Auxiliary Filtering (PAF)}, which combines the adapting model and EMA model in a primary-auxiliary structure to incorporate domain knowledge while maintaining stability. The primary filter reflects test domain knowledge to achieve strong filtering performance, while the auxiliary filter ensures stability and prevents error accumulation caused by the primary filter. The two filters collaboratively categorize and adapt to the data through complementary soft and hard filtering strategies.

On the other hand, while typical TTA methods utilize either the adapting model \cite{wang2021tent, niu2022eata, yu2025stamp, gao2024unient}, the EMA model \cite{wang2022continual, zhao2023delta}, or their ensemble for inference \cite{dobler2023rmt}, both models have limitations for inference in the OSTTA setting. Specifically, while our filtering reduces exposure to open-set data, the adapting model may still receive some open-set samples, potentially destabilizing the adapting model and impacting the EMA model. To address this, we propose \textbf{Knowledge-Integrated Prediction (KIP)}, which incorporates the source pre-trained model that remains unaffected by open-set samples. KIP combines the logits of the adapting model, EMA model, and source model, and adaptively assigns higher weights to models that exhibit greater confidence for each sample. 

To demonstrate the effectiveness of the proposed method, we evaluate our approach across various open-set TTA scenarios. We use CIFAR10/100-C and ImageNet-C \cite{deng2009imagenet, croce2021robustbench} as closed-set datasets, while SVHN \cite{netzer2011svhn}, Tiny-ImageNet \cite{le2015tinyimagenet}, Places365 \cite{zhou2017places}, and Textures \cite{cimpoi2014textures} serve as open-set datasets. Whereas prior OSTTA studies focus on either closed-set accuracy or open-set discrimination, our method achieves substantial improvements on both objectives.

\section{Primary-Auxiliary Filtering and Knowledge-Integrated Prediction}
We first present the preliminaries in \cref{sec:3_1}, followed by the motivation in \cref{motivation}. Then, we introduce our proposed methods, Primary-Auxiliary Filtering (PAF) and Knowledge-Integrated Prediction (KIP), in \cref{method1} and \cref{method2}, respectively.

\vspace{-1em}
\subsection{Preliminary}
\label{sec:3_1}
Let the source domain dataset with the label space \(\mathcal{Y}_s = \{1, \cdots, \mathcal{C}_s\}\) be denoted as \(\mathcal{D}_s = \{x_i, y_i\}_{i=1}^{N_s}\), and the test domain dataset be denoted as $\mathcal{D}_t = \{x_j, y_j\}_{j=1}^{N_t}$. If the label space of the test domain dataset \(\mathcal{Y}_t\) is the same as \(\mathcal{Y}_s\), it represents a standard closed-set TTA scenario. In contrast, open-set TTA refers to the setting where \(\mathcal{Y}_s \subset \mathcal{Y}_t\), meaning that the label set of the test domain includes additional open-set classes.

For TTA, a source pre-trained model \( f_{\theta_0} \) trained on \(\mathcal{D}_s\) is given. The model deployed in the environment encounters streaming test batch \(\bold{X}_t\) at time \( t \) for inference. Since the test batch does not have any labels, most TTA models minimizes the entropy of prediction as follows:
\begin{equation}
\min_{\theta_t} \mathcal L_{TTA}(x) = \sum_{x \in \bold X_t} H(f_{\theta_t}(x)),
\label{eq:tta}
\end{equation}
where $H(\cdot)$ denotes the entropy. However, in the OSTTA scenario, the presence of open-set data makes it undesirable to increase the reliability of all data. To address this problem, existing OSTTA studies filter data based on confidence for test-time training as follows:
\vspace{-0.5em}
\begin{equation}
\begin{split}
\min_{\theta_t} \mathcal L_{OSTTA}(x) = \sum_{x \in \bold X_t} \Big [ \mathbb{I} [\text{conf}(x) > \tau ] \cdot H(f_{\theta_t}(x))  -\alpha \cdot \mathbb{I} [\text{conf}(x) \le \tau ] \cdot H(f_{\theta_t}(x)) \Big ],
\label{eq:ostta}
\end{split}
\end{equation}

where \(\text{conf}(\cdot)\) is a function that measures the confidence of data, such as entropy or comparison with the source model, and $\tau$ is a confidence threshold. When \(\alpha = 0\), the OSTTA method disregards low-confidence data \cite{yu2025stamp, lee2023ostta, gong2024sotta}, while for \(\alpha > 0\), it maximizes the entropy of low-confidence data \cite{gao2024unient}.

\vspace{-1em}
\subsection{Motivation}

\label{motivation}

While UniEnt has demonstrated the effectiveness of entropy maximization, we found that it relies on source knowledge for filtering and incorrectly identifies many closed-set samples as open-set. 
Furthermore, we observed that incorporating domain knowledge into training is prone to error accumulation. 
To validate this observation, we conduct experiments in a continual test-time adaptation setting, where CIFAR-100-C and Textures-C are used as the closed-set and open-set datasets, respectively. The model sequentially adapts to 15 different domains, and we update only the batch normalization layers. The domain changes every 100 batches, with transition points indicated by dashed lines in \cref{fig:motivation}. ``Wrongly filtered closed-set'' in \cref{fig:unientlog} and \cref{fig:ourslog} refers to the proportion of closed-set samples that were incorrectly assigned low confidence and trained with the entropy maximization loss. 

\vspace{-1em}
\paragraph{Limitation for using fixed source model for measuring confidence.}
As mentioned earlier, existing methods rely on fixed source domain knowledge to measure $\text{conf}(\cdot)$. \Cref{fig:unientlog} shows the proportion of closed-set samples that were incorrectly assigned low confidence and trained with the entropy maximization loss. Since they fail to incorporate test domain knowledge, they assign low confidence even to many closed-set samples. As a result, more than half of the closed-set data receive lower confidence scores than open-set data and are incorrectly trained with the entropy maximization loss.

\paragraph{Using the adapting model causes error accumulation.}
One simple way to utilize test domain knowledge for confidence measurement is to use the adapting model. We applied a strategy that selects the loss based on the confidence of the adapting model. Specifically, we designed a method that determines whether to apply entropy minimization or maximization using the entropy of the adapting model $f_{\theta_t}(x)$ as follows:
\vspace{-0.3em}
\begin{equation}
\begin{split}
\mathcal L_{adapt}(x) = \sum_{x \in \mathbf X_t} &\Big [\mathbb{I} \big [ H \big( f_{\theta_t}(\tilde x) \big) < \tau \big ] \cdot H(f_{\theta_t}(\tilde x)) -  \mathbb{I} \big [ H \big( f_{\theta_t}(\tilde x) \big) \ge \tau \big ] \cdot H(f_{\theta_t}(\tilde x)) \Big ],
\label{eq:adapt}
\end{split}
\end{equation}

where $\tilde x$ is an augmented view of $x$. Following the confidence threshold used in previous methods \cite{niu2022eata}, we set $\tau = 0.4 \times \log(C)$, where $C$ denotes the number of classes. Since this method directly utilizes the knowledge of the adapting model for filtering, it can immediately reflect the characteristics of each test domain. As shown in \cref{fig:ourslog}, it achieves high filtering accuracy during the first 300 test batches. However, as more test batches accumulate, error accumulation occurs, leading to a significant drop in filtering accuracy. As a result, the model performance significantly degrades as training progresses, as shown in \cref{fig:hs}. 

\vspace{-1em}
\paragraph{EMA enables stable filtering but exhibits suboptimal performance.}
Inspired by Mean Teacher \cite{tarvainen2017mean}, which addresses a similar confirmation bias issue in semi-supervised learning, we found that the EMA model $f^\text{EMA}_{t}(x)$ obtained by aggregating the parameters of the adapting model can provide more stable filtering. For each test time step $t$, the EMA model $f^\text{EMA}_{t}(x)$ is updated as follows:

\vspace{-2em}
\begin{equation}
\begin{split}
f^\text{EMA}_{t}(x) = \beta \cdot f^\text{EMA}_{t-1} (x) + (1 - \beta) \cdot f_{\theta_t}(x),
 \end{split}
 \label{eq:ema_update}
\end{equation}
where $\beta$ is the decay rate that determines the weight of past parameters in the EMA model. Instead of using the adapting model, which causes error accumulation, we designed a strategy that utilizes the entropy of the EMA model as a confidence measure, as follows:
\begin{equation}
\begin{split}
\mathcal L_{ema}(x) = \sum_{x \in \mathbf X_t} &\Big [\mathbb{I} \big [ H \big( f^\text{EMA}_{t}(\tilde x) \big) < \tau \big ] \cdot H(f_{\theta_t}(\tilde x)) -  \mathbb{I} \big [ H \big( f^\text{EMA}_{t}(\tilde x) \big) \ge \tau \big ] \cdot H(f_{\theta_t}(\tilde x)) \Big ].
\label{eq:ema}
\end{split}
\end{equation}

The green graph in \cref{fig:ourslog} represents the filtering results of the EMA model, while the green graph in \cref{fig:hs} shows the overall performance. Since the EMA model aggregates the adapting model over multiple time steps, it reduces the rapid propagation of errors and provides more stable filtering. 

However, unlike semi-supervised learning where Mean Teacher \cite{tarvainen2017mean} is used, the EMA model in a continually shifting TTA environment accumulates knowledge from past domains. As a result, the EMA model still exhibits suboptimal filtering performance. In practice, the filtering behavior of the EMA model fluctuates significantly across different domains, and before error accumulation occurs (before 300 batches), its filtering performance is lower than that of the adapting model. To address this, we propose a method that leverages both the adapting model, which effectively reflects current domain knowledge but is prone to error accumulation, and the EMA model, which provides stable filtering. As shown in \cref{fig:ourslog} and \cref{fig:hs} (red graph), our approach maintains consistently high filtering performance across domains while preventing error accumulation over time.

\vspace{-1em}
\subsection{Primary-Auxiliary Filtering}
\label{method1}
As aforementioned, using the adapting model or the EMA model individually for filtering has distinct limitations. Therefore, we propose \textbf{Primary-Auxiliary Filtering (PAF)}, which leverages both models for filtering. PAF is a simple yet effective method. First, we define the primary filter $F_{pr}(x)$ that utilizes the adapting model $f_{\theta_t}(x)$, and identify each sample as follows:
\begin{equation}
    F_{pr}(x) = \mathbb{I} \Big[ H(f_{\theta_t}(\tilde x)) < \tau \Big].
 \label{eq:F_pr}
\end{equation}
A sample $x$ is identified as a potential open-set sample if $F_{pr}(x) = 0$. However, directly applying entropy maximization to these samples leads to error accumulation. To mitigate this issue, we perform an additional filtering step using the stable EMA model $f^\text{EMA}_t(x)$ as an auxiliary filter. Similar to $F_{pr}(x)$, the auxiliary filter $F_{aux}$ is defined as follows:
\vspace{-0.5em}
\begin{equation}
    F_{aux}(x) = \mathbb{I} \Big[  H(f^{EMA}_{t}(\tilde x)) < \tau \Big],
\label{eq:F_aux}
\end{equation}
where the only difference is that the EMA model is used instead of the adapting model. Finally, samples for which both filters output $0$ are trained with entropy maximization. Meanwhile, samples with $F_{pr}(x)=0$ but $F_{aux}(x) = 1$ are identified as potentially prone to error accumulation and are excluded from training.

Conversely, if $F_{pr}(x) = 1$, the sample $x$ is identified as a closed-set instance. However, we found that applying double filtering with $F_{aux}(\cdot)$ may lead to overly conservative training or even underfitting. Nevertheless, relying solely on $F_{pr}(\cdot)$ may still result in incorrect training. To address this, we propose a soft filtering method that softly integrates \( F_{aux}(\cdot) \) into training  for samples identified as closed-set. Specifically, we integrate the confidence from the EMA mode \( f^{EMA}_{t} \) into entropy minimization by weighting the loss as follows:
\vspace{-0.5em}
\begin{equation}
\begin{split}
w_{soft}(x) = \frac{1}{\exp [H(f^\text{EMA}_t(\tilde x)) - \tau]},
 \label{eq:w_soft}
 \end{split}
\end{equation} which assigns higher weights to high-confidence samples above the threshold \cite{niu2022eata}. 
The weight computed as above is multiplied by the entropy minimization loss. In other words, for data that $F_{pr}(\cdot)$ considers reliable, the influence on training increases as the EMA model assigns higher confidence. Conversely, if the entropy from the EMA model exceeds the threshold, a small weight is assigned to the loss. We further analyze the effects of soft and hard filtering. Finally, our proposed loss $\mathcal L_{PAF}(x)$ is defined as follows:
\begin{equation}
\begin{split}
& \min_{\theta_t} \mathcal L_{PAF}(x) = \sum_{x \in \mathbf X_t} \Big[ \mathbb{I} [F_{pr}(x) = 1] \cdot w_{soft}(x) \cdot H(f_{\theta_t}(\tilde x)) \\ &-  \alpha \cdot \mathbb{I} \big[ (F_{pr}(x) = 0) \wedge (F_{aux}(x) = 0) \big] \cdot H(f_{\theta_t}(\tilde x)) \Big].
 \label{eq:paf}
\end{split}
\end{equation}
 With $\mathcal L_{PAF}(\cdot)$, we update only the batch normalization layers of the model. On the other hand, previous OSTTA studies \cite{gao2024unient, yu2025stamp} output the open-set score of each sample to validate the open-set discrimination performance. Therefore, we output the energy score of the adapting model as the open-set score for fair comparison.
 
\vspace{-1em}
\subsection{Knowledge-Integrated Prediction}
\label{method2}
As mentioned in \cref{motivation}, the adapting model can respond quickly to distribution shifts, whereas the EMA model provides more stable and reliable predictions. Many existing TTA methods leverage either the adapting model \cite{wang2021tent, niu2022eata, yu2025stamp, gao2024unient} or the EMA model \cite{wang2022continual, zhao2023delta}. Notably, RMT \cite{dobler2023rmt} ensembles both models during inference to exploit these complementary strengths.

However, both models have limitations for inference in the OSTTA setting. Although PAF significantly improves filtering accuracy as shown in \cref{fig:ourslog}, some samples are still incorrectly filtered. This may be attributed to the fact that the model is trained to suppress the confidence of certain closed-set logits, which degrades its predictive performance on closed-set samples near the decision boundary.

Although the source model $f_{\theta_0}(\cdot)$ is not adapted to the target domain, it is important to highlight that it has never encountered open-set samples. Therefore, we propose \textbf{Knowledge-Integrated Prediction (KIP)}, which calibrates the outputs of the source, adapting, and EMA models to produce more reliable predictions. We design a per-sample weighted prediction scheme, where the weight assigned to each model’s output is determined by comparing its confidence to the average confidence across all models. If a model yields a lower confidence for a given sample, its importance in the ensemble is reduced relative to uniform averaging, and vice versa. Specifically, we define the weight $c_i(x)$ assigned to the $i$-th model as follows:
\vspace{-0.5em}
\begin{equation}
c_i(x) = \frac{1}{3} + \gamma\cdot(\max(p_i(x)) - \sum_{j=1}^3 \frac{\max(p_j(x))}{3}).
\end{equation}
Without loss of generality, we denote the probability vectors from the source, adapting, and EMA models as $p_1(x), p_2(x), p_3(x)$. The hyperparameter $\gamma$ controls the sharpness of the weights. Since the upper bound of entropy depends on the number of classes, we instead use the maximum class probability, whose upper bound is independent of the number of classes. Finally, the aggregated logit $z_{kip}(x)$ is computed as:

\begin{equation}
z_{kip}(x) = \sum_{i=1}^3 c_i(x)\cdot z_i(x),
\label{eq:kip}
\end{equation}
where $z_i(x)$ denotes the logit of the $i$-th model. Since the source model was not trained with augmentations, it operates on the original input $x$ rather than the augmented version $\tilde{x}$. 

\begin{table}[t]
\caption{Results of various methods on CIFAR10-C benchmarks with different open-set datasets. All values are averaged over 15 domains. \textbf{Bold} indicate the best results, \underline{underline} indicate the second-best, and \textcolor{red}{red} numbers represent the improvement over the previous best.}
\scriptsize
    \centering
        \resizebox{0.99\linewidth}{!}{%
        \begin{tabular}{l|ccc|ccc|ccc|ccc}
        \toprule
            \multirow{2}{*}{} & \multicolumn{3}{c|}{\textbf{SVHN-C}} & \multicolumn{3}{c|}{\textbf{TinyImageNet-C}} & \multicolumn{3}{c|}{\textbf{Places365-C}} & \multicolumn{3}{c}{\textbf{Textures-C}} \\ \cmidrule(lr){2-4} \cmidrule(lr){5-7} \cmidrule(lr){8-10} \cmidrule(lr){11-13}
                   & ACC &AUR & H-S &  ACC & AUR & H-S & ACC  &AUR & H-S & ACC  & AUR & H-S \\\midrule
Source    & 81.73 & 77.89 & 79.76 & 81.73 & 80.43 & 81.07 & 81.73 & 83.60 & 82.65 & 81.73 & 82.60 & 82.16 \\
TENT      & 78.05  & 67.64 & 72.47 & 56.91 & 62.44 & 59.55 & 49.31 & 55.83 & 52.37 & 65.92 & 64.68 & 65.29 \\
CoTTA     & \underline{85.23} & 82.91 & 84.05 & 85.88 & 79.20 & 82.40 & 85.01 & 82.71 & 83.84 & 83.17 & 79.44 & 81.26 \\
EATA      & 84.47 & 82.24 & 83.34 & 84.97 & 78.07 & 81.37 & 84.65 & 80.63 & 82.59 & 82.45 & 76.27 & 79.24 \\
RoTTA     & 85.08 & 82.97 & 84.01 & 85.12 & 80.09 & 82.53 & 85.09 & 82.33 & 83.69 & \underline{84.80} & 80.89 & 82.80 \\
SoTTA     & 69.52 & 46.91 & 56.02 & 79.98 & 63.03 & 70.50 & 79.46 & 63.23 & 70.42 & 69.97 & 49.62 & 58.06 \\
OSTTA & 84.11 & 71.98 & 77.57 & 85.29 & 71.18 & 77.60 & 84.69 & 73.09 & 78.46 & 83.07 & 66.79 & 74.05 \\
STAMP     & 82.95 & 73.27 & 77.81 & \underline{86.83} & 79.72 & 83.12 & \underline{85.65} & 80.19 & 82.83 & 81.48 & 66.09 & 72.98 \\
UniEnt    & 84.56 & 90.21 & 87.29 & 85.19 & 84.43 & 84.81 & 84.74 & 89.42 & 87.02 & 82.36 & 85.60 & 83.95 \\
UniEnt+   & 83.98 & \underline{92.94} & \underline{88.23} & 84.19 & \underline{85.53} & 84.85 & 84.24 & \underline{90.77} & \underline{87.38} & 81.21 & \underline{88.36} & \underline{84.63} 
\\\midrule
\textbf{Ours} & 
\textbf{87.49}\tiny\color{red}{+2.26} & \textbf{97.66}\tiny\color{red}{+4.72} & \textbf{92.30}\tiny\color{red}{+4.07} & \textbf{88.26}\tiny\color{red}{+1.43} & \textbf{90.61}\tiny\color{red}{+5.08} & \textbf{89.42}\tiny\color{red}{+4.57} & \textbf{88.35}\tiny\color{red}{+2.70} & \textbf{94.24}\tiny\color{red}{+3.48} & \textbf{91.20}\tiny\color{red}{+3.82} & \textbf{87.80}\tiny\color{red}{+3.00} & \textbf{97.71}\tiny\color{red}{+9.35} & \textbf{92.49}\tiny\color{red}{+7.86} \\\bottomrule
        \end{tabular}
        }

    \label{tab:cifar10}
\end{table} 

\begin{table}[t]
    \caption{Results of various methods on CIFAR100-C with different open-set datasets.}
\scriptsize
    \centering
        \resizebox{0.99\linewidth}{!}{%
        \begin{tabular}{l|ccc|ccc|ccc|ccc}
        \toprule
            \multirow{2}{*}{} & \multicolumn{3}{c|}{\textbf{SVHN-C}} & \multicolumn{3}{c|}{\textbf{TinyImageNet-C}} & \multicolumn{3}{c|}{\textbf{Places365-C}} & \multicolumn{3}{c}{\textbf{Textures-C}} \\ \cmidrule(lr){2-4} \cmidrule(lr){5-7} \cmidrule(lr){8-10} \cmidrule(lr){11-13}
         & ACC & AUR & H-S &  ACC & AUR & H-S & ACC  & AUR & H-S & ACC   & AUR & H-S\\\midrule
Source   & 53.25 & 60.55 & 56.67 & 53.25 & 68.91 & 60.08 & 53.25 & 66.92 & 59.31 & 53.25 & 62.37 & 57.45 \\
TENT     & 20.76 & 67.69 & 31.77 & 20.91 & 56.87 & 30.58 & 19.85 & 59.50 & 29.77 & 22.86 & 61.13 & 33.28 \\
CoTTA    & 55.44 & 74.07 & 63.42 & 57.61 & 68.97 & 62.78 & 55.21 & 71.94 & 62.47 & 52.13 & 65.01 & 57.86 \\
EATA     & 52.81 & 70.79 & 60.49 & 54.25 & 68.00 & 60.35 & 53.69 & 70.81 & 61.07 & 49.75 & 57.88 & 53.51 \\
RoTTA    & 59.74 & 75.53 & 66.71 & 60.00 & 70.74 & 64.93 & 60.05 & 70.14 & 64.70 & 59.27 & 72.33 & \underline{65.15} \\
SoTTA    & 39.48 & 65.74 & 49.33 & 41.52 & 61.25 & 49.49 & 43.20 & 60.01 & 50.24 & 52.31 & 57.32 & 54.70 \\
OSTTA    & \underline{60.24} & 75.68 & 67.08 & 61.60 & 71.20 & 66.05 & 60.35 & 72.72 & 65.96 & \underline{59.57} & 66.43 & 62.81 \\
STAMP    & 59.51 & \underline{92.61} & \underline{72.46} & \textbf{64.25} & 73.16 & \underline{68.42} & \underline{62.89} & 74.23 & \underline{68.09} & 54.32 & 39.79 & 45.93 \\
UniEnt   & 59.24 & 88.91 & 71.10 & 60.34 & 73.87 & 66.42 & 59.11 & 78.10 & 67.29 & 57.02 & 73.91 & 64.38 \\
UniEnt+  & 58.74 & 91.03 & 71.40 & 60.24 & \underline{74.21} & 66.50 & 58.90 & \underline{78.64} & 67.35 & 56.58 & \underline{74.55} & 64.33 

\\\midrule
\textbf{Ours} & 
\textbf{62.59}\tiny\color{red}{+2.35} & \textbf{97.61}\tiny\color{red}{+5.00}& \textbf{76.27}\tiny\color{red}{+3.81}& \underline{63.79}\tiny\color{red}{-0.46}& \textbf{84.16}\tiny\color{red}{+9.95}& \textbf{72.57}\tiny\color{red}{+4.15}& \textbf{62.94}\tiny\color{red}{+0.05}& \textbf{85.25}\tiny\color{red}{+5.98}& \textbf{72.42}\tiny\color{red}{+4.33}& \textbf{62.81}\tiny\color{red}{+3.24}& \textbf{94.54}\tiny\color{red}{+19.99}& \textbf{75.48}\tiny\color{red}{+10.33}\\\bottomrule
        \end{tabular}
        }

    \label{tab:cifar100}
\vspace{-1.2em}
\end{table}

\section{Experiments}
\subsection{Experiments Setup}
\paragraph{Setting and Datasets.}
Following previous studies, we select three types of corrupted datasets as closed-set datasets. CIFAR10-C \cite{croce2021robustbench} is derived from the CIFAR10 test data with 15 types of corruption, containing 10 classes. Similarly, CIFAR100-C and ImageNet-C \cite{croce2021robustbench} are based on the CIFAR100 and ImageNet test datasets. Each corruption has an intensity level ranging from 1 to 5, and we use the highest level. For CIFAR benchmarks, we use SVHN-C \cite{netzer2011svhn}, Tiny-ImageNet-C \cite{le2015tinyimagenet}, Places365-C \cite{zhou2017places}, and Textures-C \cite{cimpoi2014textures} as open-set datasets. For ImageNet-C \cite{deng2009imagenet}, we use Places365-C and Textures-C. All open-set data were resized to match the size of the closed-set data and had the same types of corruption applied as the closed-set. Following prior studies \cite{yu2025stamp, gao2024unient, lee2023ostta}, we assume a continual OSTTA environment, where domains change continuously, and open-set and closed-set data are mixed at a 1:1 ratio. The model sequentially adapts to each of the 15 corruption domains, and no reset is performed throughout the adaptation process.

All models are provided by RobustBench \cite{croce2021robustbench} and are pre-trained on a clean dataset using AugMix augmentation. For CIFAR benchmarks, the WideResNet-40-2 \cite{zagoruyko2016wide} architecture is used. During TTA, a batch size of 200 and the Adam \cite{kingma2014adam} optimizer with a learning rate of 0.001 are utilized. For ImageNet benchmarks, the ResNet-50 \cite{he2016deep} architecture is used, with a batch size of 64 and the SGD optimizer with a learning rate of 0.00025. Note that SoTTA and STAMP are implemented with the SAM optimizer \cite{foret2020sam} as in the original paper, and their learning rates are maintained as specified in the respective studies. 

\vspace{-1em}
\paragraph{Evaluation protocols.}
We evaluate both the classification capability on closed-set data and the discrimination ability for open-set data. The classification performance on closed-set data is measured using classification accuracy \textbf{(ACC)}. For open-set discrimination ability, we use the area under the ROC curve \textbf{(AUR)}, calculated based on the open-set scores generated by the model for each data point. Following prior studies, the energy score is used for all methods. To combine both metrics, we use the harmonic mean \textbf{(H-S)} of ACC and AUR. We compare our method with nine TTA methods, including open-set TTA approaches. The hyperparameters for each method are set as closely as possible to those in the original papers. 

\vspace{-1.3em}
\paragraph{Implementation details.}
We use simple random crop and horizontal flip augmentations. The loss balancing hyperparameter $\alpha$ is set to 2.0 and 0.7 for CIFAR and ImageNet, respectively.  Following Mean Teacher \cite{tarvainen2017mean}, we set $\beta$ as 0.999. $\gamma$ is set to 0.1. While several OSTTA methods employ dataset-specific thresholds, we set $\tau$ to $0.4 \times \log(C)$ across all settings, where $C$ is the number of closed-set classes, following \cite{niu2022eata}. We later show that our method is robust to hyperparameter selection. 

\begin{table}[t]
\parbox[t]{0.58\linewidth}{
\caption{Results of various methods on ImageNet-C benchmarks with different open-set datasets.}
\scriptsize
\centering
    \resizebox{0.90\linewidth}{!}{%
        \begin{tabular}{l|ccc|ccc}
            \toprule
            \multirow{2}{*}{} & \multicolumn{3}{c|}{\textbf{Places365-C}}& \multicolumn{3}{c}{\textbf{Textures-C}} \\ \cmidrule(r){2-4} \cmidrule(r){5-7} 
          &  ACC  & AUR & H-S & ACC   & AUR & H-S \\\midrule
Source   & 28.22 & 67.86 & 39.86   & 28.22 & 71.55 & 40.48  \\
TENT     & 41.28 & 65.91 & 50.77   & 35.12 & 47.30 & 40.31  \\
CoTTA    & 41.41 & 72.64 & 52.75   & 40.81 & 70.05 & 51.57 \\
EATA     & \underline{46.99} & 78.95 & 58.91   & 43.42 & 65.62 & 52.26  \\
RoTTA    & 44.15 & 72.14 & 54.78   & 45.13 & 77.60 & 57.07  \\
SoTTA    & 2.66  & 50.49 & 5.05    & 2.59  & 40.46 & 4.87   \\
OSTTA    & 47.05 & 75.62 & 58.01   & 45.03 & 61.73 & 52.07 \\
STAMP    & 46.88 & \underline{80.11} & \underline{59.15}  & \underline{47.49} & \underline{80.59} & \underline{59.76}  \\
UniEnt   & 46.16 & 78.81 & 58.22   & 44.46 & 64.01 & 52.47  \\
UniEnt+  & 45.82 & 78.75 & 57.93   & 44.23 & 63.50 & 52.14 \\\midrule

\textbf{Ours} & \textbf{48.22}\tiny\color{red}{+1.17} & \textbf{84.18}\tiny\color{red}{+4.07} & \textbf{61.32}\tiny\color{red}{+2.17} & \textbf{47.96}\tiny\color{red}{+0.47} & \textbf{82.91}\tiny\color{red}{+2.32} & \textbf{60.77}\tiny\color{red}{+1.01} \\\bottomrule
        \end{tabular}}

    \label{tab:imagenet}}
\hspace{0.05\linewidth}
\parbox[t]{0.33\linewidth}{
\caption{Ablation study of $F_{pr}$, $F_{aux}$ and KIP.}
    \centering
\scriptsize
\setlength{\tabcolsep}{0.1cm}         
\renewcommand{\arraystretch}{0.8}       
\begin{tabular}{ccc|ccc}
\toprule \rule{0pt}{1.0EM}
$F_{pr}$ & $F_{aux}$ & \textit{KIP} & ACC & AUR & H-S \\ \midrule
\xmark & \xmark & \xmark  & 20.76 & 67.69 & 31.77 \\\cmidrule(){1-6} 
\cmark & \xmark & \xmark  & 26.27 & 82.98 & 39.91 \\
\xmark & \cmark & \xmark  & 55.83 & 97.11 & 70.90  \\ 
\cmark & \cmark & \xmark  & 58.27 & 97.61 & 72.98 \\  \cmidrule(){1-6} 
\cmark & \cmark & \cmark  & 62.59 & 97.61 & 76.27 \\
\bottomrule
\end{tabular}

\label{tab:ab1}
}\vspace{-1em}
\end{table} 

\vspace{-1em}
\subsection{Results}
\paragraph{CIFAR benchmarks.}
Table \ref{tab:cifar10} shows the results of adding various open-set datasets to an environment where CIFAR10-C is used as the closed-set dataset. Existing methods achieve strong performance in only one of the two metrics. Generally, RoTTA and STAMP, which focus on stable learning for closed-set data, exhibit high ACC but lower AUROC. In contrast, UniEnt and UniEnt+ tend to have relatively lower ACC but higher AUROC. Nevertheless, our method consistently improves performance across all scenarios. In particular, when Textures-C was used as the open-set dataset, our method demonstrates a performance increase of 3.00\%p in ACC and 9.35\%p in AUROC.  

Table \ref{tab:cifar100} presents the evaluation results where CIFAR100-C serves as the closed-set dataset. Similar to CIFAR10-C, methods focused on stable learning exhibit high ACC but lower AUROC, while UniEnt shows relatively lower ACC but higher AUROC. On the other hand, our method achieves performance improvements across most cases. Notably, when Textures-C is used as the open-set dataset, our approach achieves 19.99\%p higher AUROC than the best existing method. This indicates that our method effectively handles open-set data and allows the model to achieve strong discriminative capability.

\vspace{-1em}
\paragraph{ImageNet benchmarks.}
Table \ref{tab:imagenet} presents the results of adding various open-set datasets to an environment where ImageNet-C is used as the closed-set dataset. ImageNet-C poses a more challenging task due to its 1000 classes and larger image size. As a result, UniEnt achieves relatively low AUROC, even when entropy maximization is applied to low-confidence samples. However, our method still achieves superior performance. Notably, in the environment where Places365-C is the open-set dataset, our approach significantly improves the H-score.
\vspace{-0.5em}
\subsection{Further Analysis}
\vspace{-0.5em}
\paragraph{Ablation study for $F_{pr}$, $F_{aux}$, and KIP.}
Table \ref{tab:ab1} shows the ablation study of the main components of our method, using CIFAR100-C as the closed-set and SVHN-C as the open-set dataset. Without any filtering, the model minimizes the entropy of open-set samples, leading to severe performance degradation. Using only $F_{pr}$ leads to confirmation bias, while $F_{aux}$ prevents error accumulation but fails to incorporate target-domain knowledge, resulting in lower accuracy. In contrast, combining both as a primary-auxiliary filter improves both metrics. Finally, integrating all components yields further improvements in both metrics.

\begin{figure}[t]
\centering
\begin{minipage}[t]{0.4\linewidth}
  \vspace{0pt}
  \centering
  \includegraphics[width=\linewidth]{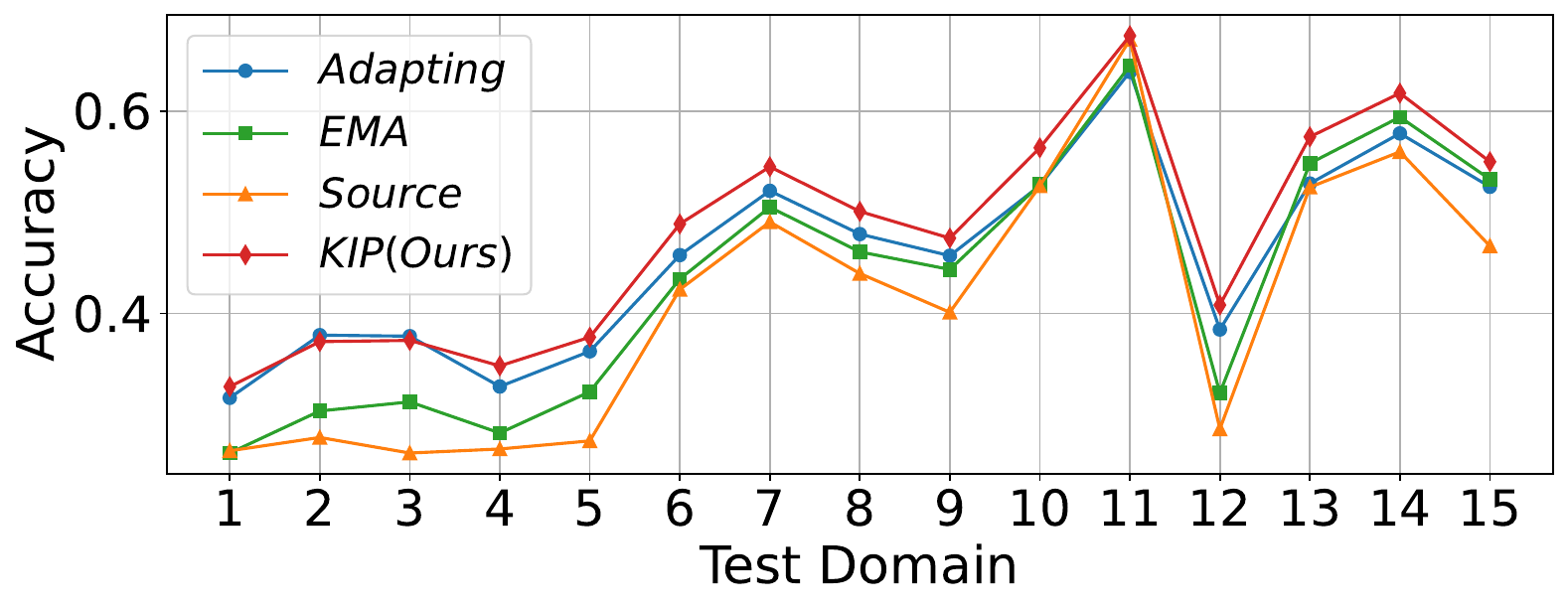}
  \vspace{-2em}
  \caption{Accuracy of KIP and independent models used in KIP.}
  \label{fig:kip_log}
  
\end{minipage}
\hspace{0.01\linewidth}
\begin{minipage}[t]{0.53\linewidth}
  \vspace{0pt} 
  \centering

\begin{tabular}{cc} \scriptsize
(a) CIFAR100-C + SVHN-C  &\scriptsize (b) ImageNet-C + Textures-C \\

\scriptsize
 \begin{tabular}{c|cc}
      \toprule & \multicolumn{2}{c}{\textbf{Ent.Max}} \\ \cline{2-3}
      \textbf{Ent.Min} & Soft & Hard \\ \midrule
      Soft & 75.92 & \textbf{76.27}\\
      Hard & 73.85 & 75.42 \\ \bottomrule
    \end{tabular}\vspace{0.1cm} &
    \scriptsize
     \begin{tabular}{c|cc}
      \toprule & \multicolumn{2}{c}{\textbf{Ent.Max}} \\ \cline{2-3}
      \textbf{Ent.Min} & Soft & Hard \\ \midrule
      Soft & 55.52 & \textbf{60.77}\\
      Hard & 47.04 & 59.49\\ \bottomrule
    \end{tabular}\vspace{0.05cm}  \\
\end{tabular}

  \captionof{table}{Ablation study on soft and hard filtering}
  \centering
  \label{tab:soft_hard}
\end{minipage}

\end{figure}

\begin{figure}[t]
  \centering
  \begin{minipage}[t]{0.48\linewidth}
    \vspace{0pt}
    \centering
    \includegraphics[width=\linewidth]{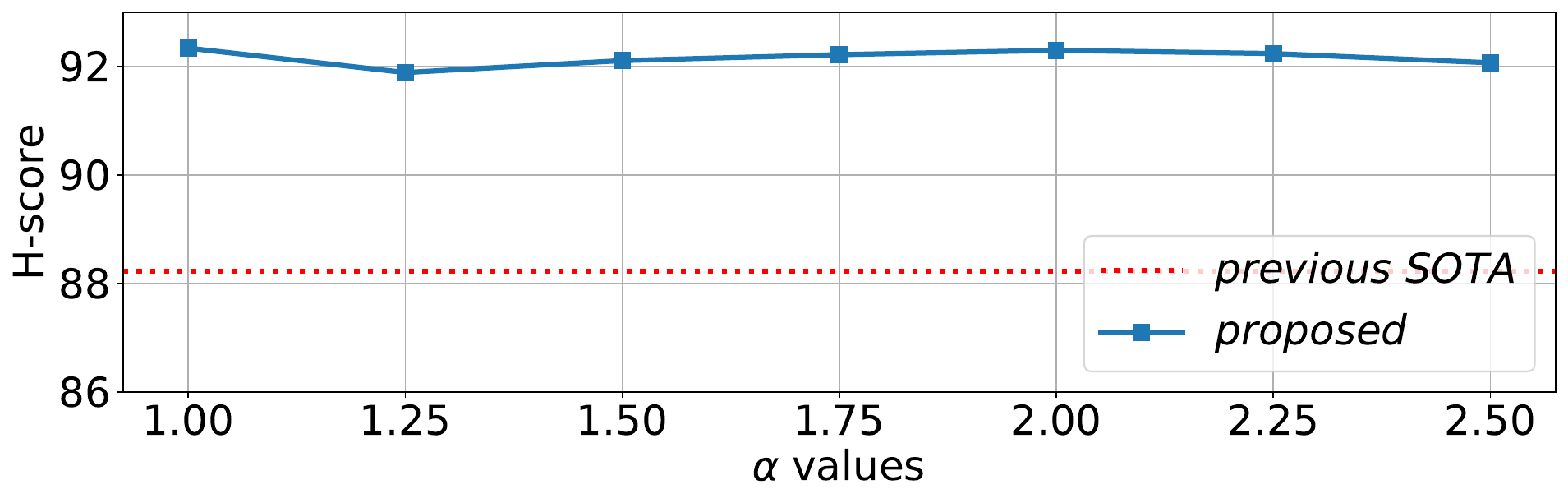}
    \vspace{-2em}
    \caption{H-score with various loss-weighting hyperparameters $\alpha$}
    \label{fig:alpha_ab}
  \end{minipage}
  \hfill
  \begin{minipage}[t]{0.48\linewidth}
    \vspace{0pt}
    \centering

    \includegraphics[width=\linewidth]{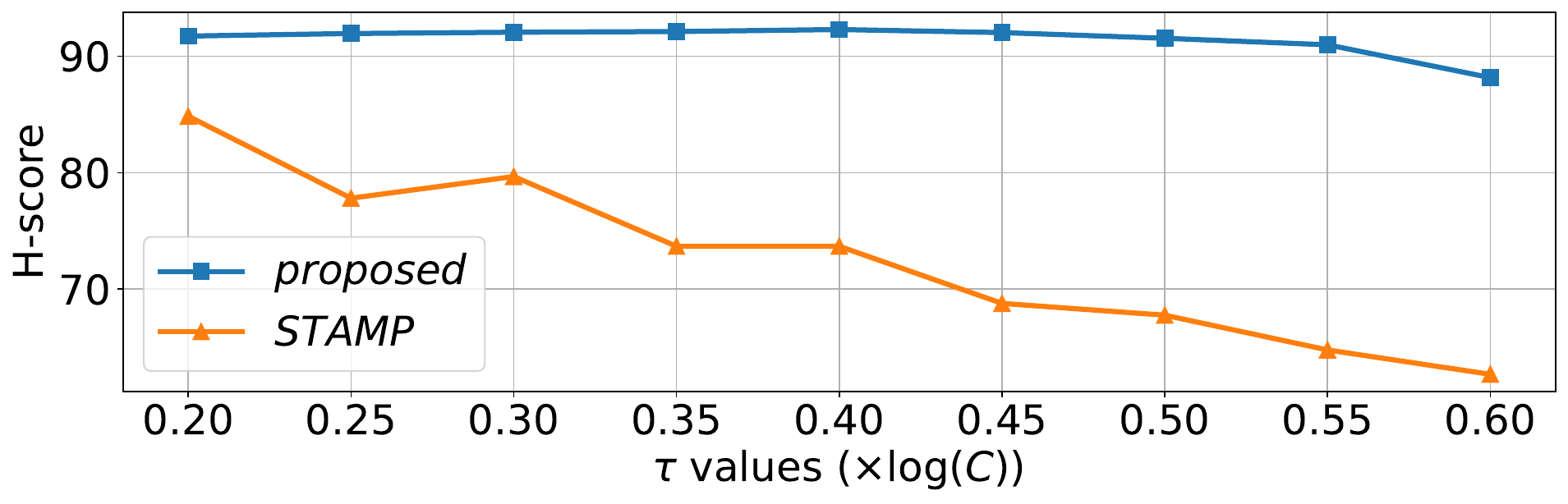}
    \vspace{-2em}
    \caption{H-score with various entropy thresholds $\tau$}
    \label{fig:tau_ab}
  \end{minipage}
\vspace{-1.5em}
\end{figure}

\vspace{-1em}
\paragraph{Soft filtering and hard filtering.}
\Cref{tab:soft_hard} presents the H-score on different filtering combinations. The best performance is achieved by applying soft filtering for entropy minimization and hard filtering for entropy maximization. Not rejecting samples during entropy maximization significantly degrades performance, while using hard filtering during entropy minimization leads to overly conservative learning and reduced accuracy.
 
\vspace{-1em}
\paragraph{Ablation study of models used in KIP.}
\Cref{fig:kip_log} presents the accuracy of three models across different domains, along with the performance of KIP. ImageNet-C and Textures-C serve as the closed-set and open-set datasets, respectively. When each model is used independently, the performance remains suboptimal. In particular, learning from the open-set causes the performance of the adapting model to temporarily drop below that of the source or EMA model. In contrast, KIP effectively calibrates unstable knowledge and achieves a significant improvement. A detailed ablation and effectiveness analysis of KIP is presented in the appendix.
\vspace{-1em}
\paragraph{Hyperparameter sensitivity.}
\Cref{fig:alpha_ab,fig:tau_ab} present experiments on loss weighting hyperparameters $\alpha$ and threshold $\tau$. CIFAR10-C and SVHN-C are used as the closed-set and open-set datasets, respectively. Our method consistently achieves higher H-scores than existing approaches across all settings and remains robust to the choice of both hyperparameters. 
\vspace{-1em}

\section{Conclusion}
In this study, we propose PAF, which substantially mitigates error accumulation during sample filtering in OSTTA. We also introduce KIP, a simple and effective strategy to enhance inference in OSTTA. Unlike prior approaches that rely on extensive data augmentations \cite{yu2025stamp} or require full model fine-tuning \cite{wang2022continual, yuan2023robust}, our method is efficient, requiring only a few forward passes and updating only batch normalization layers. Moreover, unlike replay buffer-based methods \cite{gong2024sotta, yu2025stamp, yuan2023robust} that store test samples during inference, our approach operates without any external data buffer, making it more suitable for privacy-sensitive applications. Our method achieves strong performance on the CIFAR and ImageNet benchmarks.

\section*{Acknowledgments}
This work was supported by Institute of Information \& communications Technology Planning \& Evaluation(IITP) grant funded by the Korea government(MSIT) (RS-2019-II190421, AI Graduate School Support Program(Sungkyunkwan University), 20\%), (No.2022-0-01045, Self-directed Multi-modal Intelligence for solving unknown, open domain problems, 20\%),  (No.1711195788, Development of Flexible SW/HW Conjunctive Solution for on-edge self-supervised learning, 20\%), (No.RS-2025-25442569, AI Star Fellowship Support Program (Sungkyunkwan Univ.), 20\%) and (No.RS-2024-00360227, Developing Multimodal Generative AI Talent for Industrial Convergence, 20\%).

\clearpage
\setcounter{page}{1}
\setcounter{section}{0}

\renewcommand{\thesection}{\Alph{section}}
\section{Related Works}
\paragraph{Test-Time Adaptation}
Test-time adaptation enables a trained model to adapt to new domains during testing, aiming to mitigate performance degradation when the training and test domains differ \cite{chen2022adacon, sinha2023test, dobler2023rmt, liang2024survey}. This assumes an environment where unlabeled test data is continuously streamed, and the model learns the test domain during inference. Most TTA algorithms are designed to improve the reliability of the model’s predictions \cite{wang2021tent, gong2022note,  niu2022eata}. For instance, Tent \cite{wang2021tent} optimizes the model at test time by minimizing the entropy of the test data. Subsequently, CoTTA \cite{wang2022continual} introduced a practical scenario where domains shift continuously, and many recent TTA studies have adopted the setting of CoTTA for performance evaluation \cite{gan2023decorate, liu2023vida, gao2024unient}.

\paragraph{Various Scenarios in Test-Time Adaptation}
TTA research is also advancing by considering more challenging test scenarios. For example, some studies explore environments where the class distribution at test time is non-iid (non-independent and identically distributed) \cite{gong2022note, yuan2023robust, tomar2024mixing}. This causes the model to learn biased knowledge toward certain classes, significantly reducing generalization performance and accumulating errors for other classes. To prevent error accumulation in such environments, they use a strategy such as a memory buffer \cite{gong2022note} or stable batch normalization \cite{kang2025membn}. 

On the other hand, some recent studies have considered open-set TTA \cite{gao2024unient, lee2023ostta, gong2024sotta, yu2025stamp}. OSTTA \cite{lee2023ostta} avoids training on samples with confidence levels lower than those of the source model. SoTTA \cite{gong2024sotta} and STAMP \cite{yu2025stamp} address error accumulation by placing reliable samples in a memory buffer and repeatedly training on them, thereby preventing errors associated with open-set data. UniEnt \cite{gao2024unient} attempts to learn the distribution of open-set data. UniEnt leverages the knowledge of a fixed source model to evaluate data confidence, clustering data with high and low confidence using a Gaussian Mixture Model. It minimizes entropy for high-confidence data and maximizes entropy for low-confidence data. Nevertheless, UniEnt filters data using a fixed source model, which restricts its ability to effectively leverage domain knowledge and leads to the misidentification of a number of low-confidence closed-set samples as open-set.

\section{Pseudo Code}
For a clearer understanding of the proposed methods, we provide the algorithms for Primary-Auxiliary Filtering and Knowledge-Integrated Prediction as \cref{alg:algorithm1}.
\begin{algorithm}[t]

\caption{Primary-Auxiliary Filtering and Knowledge-Integrated Prediction}
\label{alg:algorithm1}

\textbf{Input:} Source pre-trained model $f_{\theta_0}$, test domain dataset $\mathcal D_t = \{x_i, y_i\}_{i=1}^{N_t}$, batch size $B = |\mathcal B_t|$

\medskip
\For {$t \leftarrow1$ \textbf{to} T}{
    \For {$x_i \in \mathcal B_t$}{
    \textcolor{teal}{\it // \textit{Primary filter}}
    
    Compute $F_{pr}(x_i)$ via \cref{eq:F_pr}
    
    \If {$F_{pr}(x_i) = 1$}{
    \textcolor{teal}{\it // \textit{Soft filtering with auxiliary filter}}

    Compute $w_{soft}(x_i)$ via \cref{eq:w_soft}}

    \If {$F_{pr}(x_i) = 0$}{
    \textcolor{teal}{\it // \textit{Hard filtering with auxiliary filter}}

    Compute $F_{aux}(x_i)$ via \cref{eq:F_aux}}
    }
    Compute $\sum _{i=1}^B\mathcal L_{PAF}(x_i) / B$ via \cref{eq:paf}
    
    Update $\theta_t$ 
    
    Compute $z_{kip}(x_i)$ via \cref{eq:kip}
    
    Update $f^{EMA}_t$ via \cref{eq:ema_update}

    \Return {The predictions $z_{kip}(x_i)$}
    
    }
\end{algorithm}

\section{Inference Latency}
\begin{table}
\caption{Inference latency per batch across different TTA methods}
    \centering
    \footnotesize
    \resizebox{0.99\linewidth}{!}{%
\begin{tabular}{c|cccccc|ccc}
\toprule \rule{0pt}{1.0EM}
 
                       & TENT & EATA & OSTTA & SoTTA & UniEnt & \textbf{Ours} & STAMP & RoTTA & CoTTA \\
\midrule
\textbf{Inference time (s)}  & 0.024 & 0.055 & 0.056 & 0.065 & 0.072 & 0.083 & 0.626 & 0.784 & 0.925\\\midrule
\textbf{H-score}                     & 31.77& 60.49 & 67.08 & 49.33 & 71.10 & \textbf{76.21} & \underline{72.46} & 66.71 & 63.42 \\
\bottomrule
\end{tabular}}
\label{tab:latency}
\end{table}
TTA often assumes real-time constraints, which makes inference latency a critical consideration. \Cref{tab:latency} reports the per-batch inference latency under the CIFAR100-C with SVHN-C setting. Previous state-of-the-art method, STAMP, utilizes 32 augmented views during inference, which leads to strong performance at the cost of significantly higher latency. In contrast, our method achieves the highest performance while maintaining inference time comparable to other efficient baselines.

\section{Further Analysis of KIP}
\begin{figure}[t]
\centering
\centerline{\includegraphics[width=.8\linewidth]{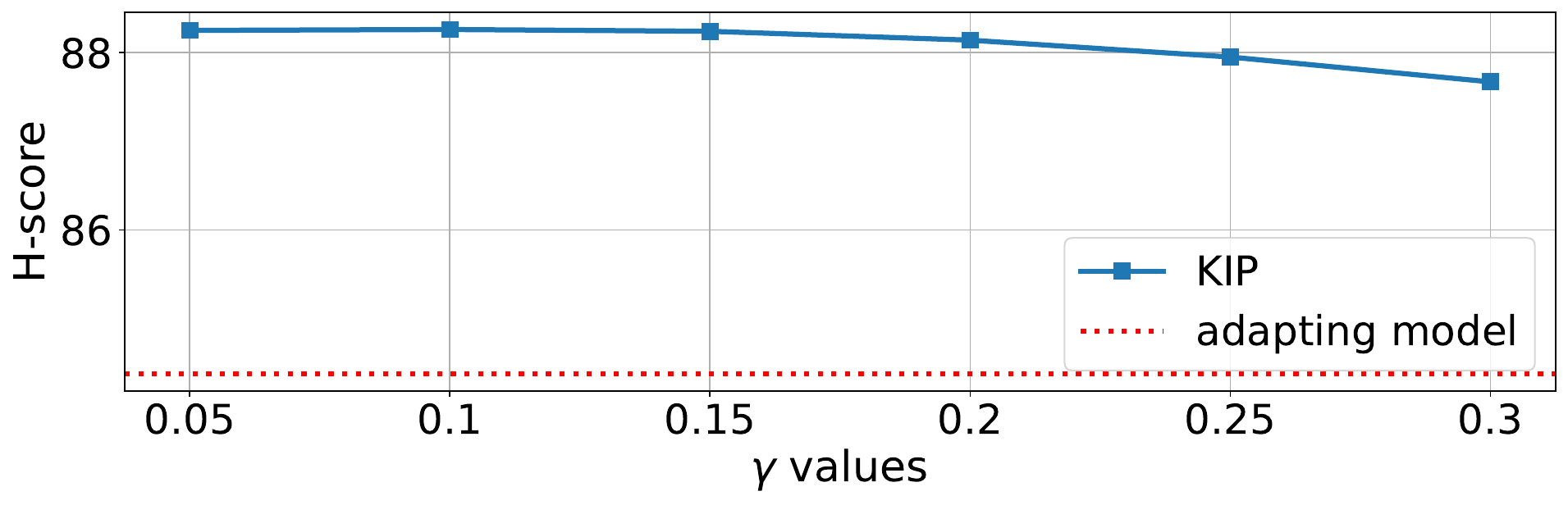}}
\caption{Accuracy with various $\gamma$.}
\label{fig:gamma}
\end{figure}
\vspace{-1em}
\begin{figure}[t]
\centering
\centerline{\includegraphics[width=.8\linewidth]{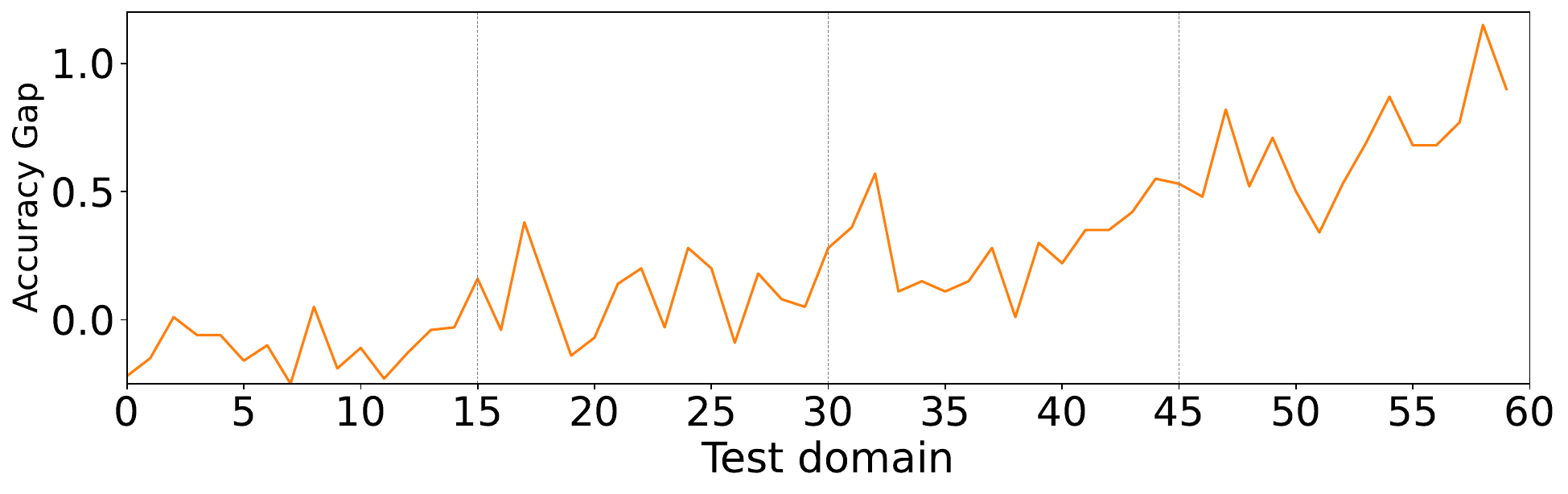}}
\caption{Accuracy gap between simple logit averaging and KIP under long-term adaptation.}
\label{fig:acc_diff}
\end{figure}

\Cref{fig:gamma} presents the performance across different values of $\gamma$, where Textures-C is used as the open-set dataset. The model shows stable performance across different values of $\gamma$, since the ensemble already tends to reduce the influence of low-confidence predictions. 

Nevertheless, adaptively weighting the logits from each model based on confidence remains beneficial. \Cref{fig:acc_diff} compares the performance of simple logit averaging and KIP in a more challenging setting with prolonged exposure to open-set data (60 domain shifts). While the two methods perform similarly in the early stages of adaptation, KIP achieves up to a 1\%p improvement as exposure to open-set data increases. This suggests that, in the later stages of adaptation, the predictions of the main model become noisier, and confidence-based weighting becomes more effective than naively averaging the logits of the three models.

\section{Performance under various EMA Decay Rates}
\begin{figure}[ht]
\centering
\centerline{\includegraphics[width=.66\linewidth]{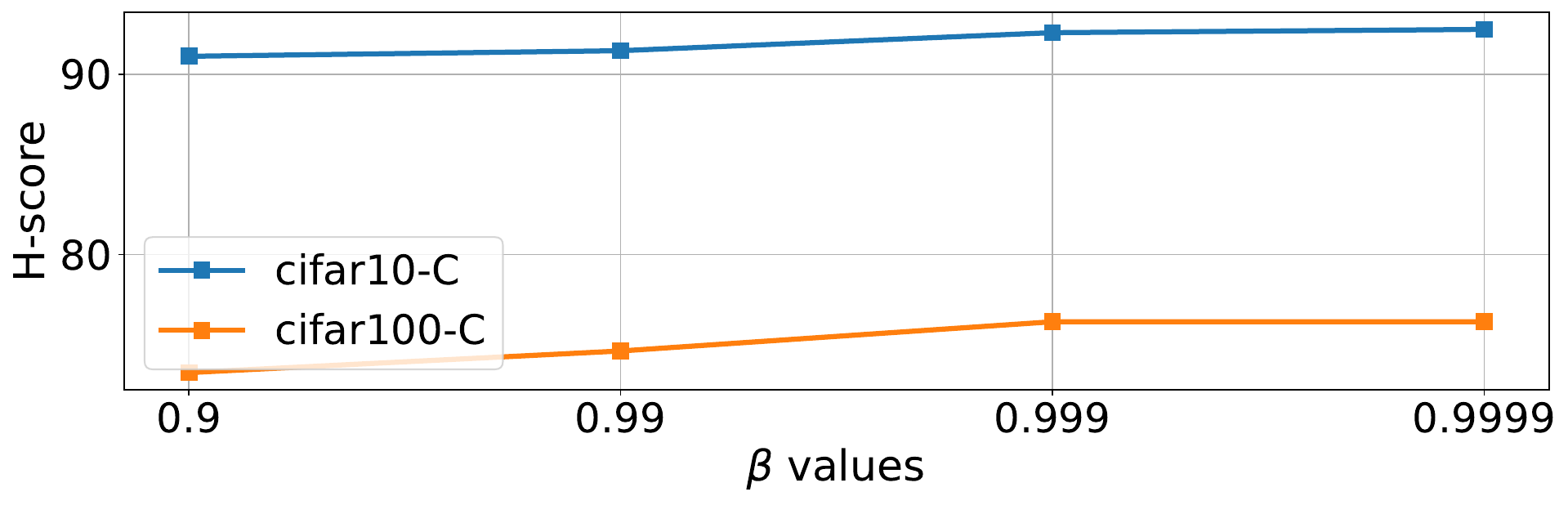}}
\caption{H-score with various EMA decay rate $\beta$.}
\label{fig:beta_ab}
\end{figure}

\Cref{fig:beta_ab} presents the performance across different EMA decay rates \( \beta \). SVHN-C is used as the open-set dataset. When \( \beta \) is set to 0.999 or higher, the model maintains stable performance. However, a smaller \( \beta \) leads to poor performance, as it heavily reflects the adapting model in the EMA model, which contradicts our motivation to prevent error accumulation.

\section{Performance under Smaller Batch Sizes}
To evaluate the robustness of our method under smaller batch sizes, we conduct experiments with reduced batch sizes of 150, 100, and 50. We compared our approach against memory-free TTA methods. In this experiment, CIFAR10-C and CIFAR100-C are used as closed-set datasets, while SVHN-C is used as the open-set dataset. 

Table \ref{tab:sup2} presents the results of the memory-free TTA methods, with smaller batch sizes. Some TTA methods exhibit significant performance degradation with smaller batch sizes. On the other hand, our method effectively preserves the model's knowledge by combining the knowledge of multiple models. Therefore, our method consistently shows high accuracy and AUROC even with smaller batch sizes. 
\vspace{-1em}
\section{Performance under Various Open-Set Data Ratios}
Following existing approaches \cite{lee2023ostta, gao2024unient}, we conducted our main experiments by setting the ratio of open-set to closed-set data at 1:1. However, in real-world scenarios, this ratio is not always balanced. To evaluate the robustness of our method under various open-set data ratios, we vary the ratio of open-set to closed-set data in our experiments. We considered cases where open-set data was less included (25\%, 50\%, 75\%) or more included (125\%, 150\%, 175\%) compared to closed-set data. For these experiments, CIFAR10-C and CIFAR100-C are used as closed-set datasets, while SVHN-C is used as the open-set dataset.  

Table \ref{tab:sup1} presents the results of experiments across various open-set scenarios with differing ratios. Some methods exhibit unstable performance under varying open-set ratios. For instance, STAMP demonstrates a significant drop in AUROC when the proportion of open-set data is low. In contrast, our method consistently shows the highest AUROC and accuracy compared to other methods, across all open-set ratios.

\begin{table}[t]
\small
\centering
\caption{Results of various methods on CIFAR benchmarks with different batch size. SVHN-C is used as the open-set dataset for these experiments. All values are averaged across 15 domains. \textbf{Bold} numbers indicate the best results, and \textcolor{red}{red} numbers represent the difference between our method and the previous best.}
\setlength{\tabcolsep}{4pt} 
\renewcommand{\arraystretch}{1.1} 
\resizebox{\textwidth}{!}{%
\begin{tabular}{ll|ccc|ccc|ccc|ccc|}
\toprule
\multirow{2}{*}{\rotatebox{90}{}} & \textbf{Batch size} & \multicolumn{3}{c|}{\textbf{50}} & \multicolumn{3}{c|}{\textbf{100}} & \multicolumn{3}{c|}{\textbf{150}} & \multicolumn{3}{c|}{\textbf{200}}\\ 
\cmidrule(lr){3-5} \cmidrule(lr){6-8} \cmidrule(lr){9-11} \cmidrule(lr){12-14}
 &  & ACC & AUR & H-S & ACC & AUR & H-S & ACC & AUR & H-S & ACC & AUR & H-S \\ 
\midrule
\multirow{8}{*}{\rotatebox[origin=c]{90}{\textbf{CIFAR10-C}}}
& Source  & 81.73 & 77.89 & 79.76 & 81.73 & 77.89 & 79.76 & 81.73 & 77.89 & 79.76 & 81.73 & 77.89 & 79.76 \\
& TENT    & 31.63 & 62.38 & 41.98 & 63.44 & 68.77 & 66.00 & 68.66 & 71.14 & 69.88 & 78.05 & 67.64 & 72.47 \\
& CoTTA   & 81.06 & 80.09 & 80.57 & 84.38 & 85.19 & 84.78 & 85.46 & 86.00 & 85.73 & 85.23 & 82.91 & 84.05 \\
& EATA    & 84.00 &	79.26 & 81.56 & 84.53 & 82.35 & 83.43 & 84.32 & 80.71 & 82.48 & 84.47 & 82.24 & 83.34 \\
& OSTTA   & 83.81 & 74.22 & 78.72 & 83.81 & 74.22 & 78.72 & 83.95 & 72.76 & 77.96 & 84.11 & 71.98 & 77.57 \\
& UniEnt  & 83.57 & 89.91 & 86.62 & 84.20 & 90.28 & 87.12 & 84.52 & 90.24 & 87.12 & 84.56 & 90.21 & 87.29 \\
& UniEnt+ & 82.59 & 92.40 & 87.22 & 83.49 & 92.73 & 87.87 & 83.90 & 92.92 & 88.18 & 83.98 & 92.94 & 88.23 \\
\cmidrule(lr){2-14}
& \textbf{Ours} 
& \textbf{84.09}\tiny\color{red}{+0.09} & \textbf{97.92}\tiny\color{red}{+5.52} & \textbf{90.47}\tiny\color{red}{+3.25} 
& \textbf{86.38}\tiny\color{red}{+1.85} & \textbf{97.93}\tiny\color{red}{+5.20} & \textbf{91.79}\tiny\color{red}{+3.92} 
& \textbf{87.39}\tiny\color{red}{+1.93} & \textbf{97.88}\tiny\color{red}{+4.96} & \textbf{92.34}\tiny\color{red}{+4.16} 
& \textbf{87.49}\tiny\color{red}{+2.26} & \textbf{97.66}\tiny\color{red}{+4.72} & \textbf{92.30}\tiny\color{red}{+4.07} \\
\bottomrule
\multirow{8}{*}{\rotatebox[origin=c]{90}{\textbf{CIFAR100-C}}}
& Source  & 53.25 & 60.55 & 56.67 & 53.25 & 60.55 & 56.67 & 53.25 & 60.55 & 56.67 & 53.25 & 60.55 & 56.67 \\
& TENT    & 8.61  & 69.01 & 15.31 & 12.69 & 66.78 & 21.33 & 17.49 & 68.74 & 27.89 & 20.76 & 67.69 & 31.77 \\
& CoTTA   & 47.58 & 71.66 & 57.19 & 54.18 & 75.70 & 63.16 & 56.00 & 76.41 & 64.63 & 55.44 & 74.07 & 63.42 \\
& EATA    & 41.25 & 61.13 & 49.26 & 44.69 & 64.14 & 52.68 & 49.58 & 66.88 & 56.95 & 52.81 & 70.79 & 60.49 \\
& OSTTA   & 57.92 & 75.06 & 65.39 & 59.26 & 74.94 & 66.18 & 59.69 & 75.03 & 66.49 & 60.24 & 75.68 & 67.08 \\
& UniEnt  & 57.32 & 89.38 & 69.85 & 58.60 & 89.30 & 70.76 & 58.91 & 89.31 & 70.99 & 59.24 & 88.91 & 71.10 \\
& UniEnt+ & 56.79 & 90.89 & 69.90 & 57.97 & 91.02 & 70.83 & 58.53 & 91.19 & 71.30 & 58.74 & 91.03 & 71.40 \\
\cmidrule(lr){2-14}
& \textbf{Ours} 
& \textbf{59.19}\tiny\color{red}{+1.27} & \textbf{97.06}\tiny\color{red}{+6.17} & \textbf{73.54}\tiny\color{red}{+3.64} 
& \textbf{61.94}\tiny\color{red}{+2.68} & \textbf{97.95}\tiny\color{red}{+6.93} & \textbf{75.89}\tiny\color{red}{+5.06} 
& \textbf{62.49}\tiny\color{red}{+2.80} & \textbf{97.63}\tiny\color{red}{+6.44} & \textbf{76.21}\tiny\color{red}{+4.91} 
& \textbf{62.59}\tiny\color{red}{+2.35} & \textbf{97.61}\tiny\color{red}{+5.00} & \textbf{76.27}\tiny\color{red}{+3.81} \\
\bottomrule
\end{tabular}}

\label{tab:sup2}
\end{table}

\begin{table}[b]
\small
\centering
\caption{Results of various methods on CIFAR benchmarks with different open-set ratios. SVHN-C is used as the open-set dataset for these experiments. All values are averaged across 15 domains. \textbf{Bold} numbers indicate the best results, and \textcolor{red}{red} numbers represent the difference between our method and the previous best.}
\setlength{\tabcolsep}{4pt}
\renewcommand{\arraystretch}{1.1}
\resizebox{\textwidth}{!}{%
\begin{tabular}{ll|ccc|ccc|ccc|ccc|ccc|ccc|}
\toprule
\multirow{2}{*}{\rotatebox{90}{}} & \textbf{Open-set ratio} & \multicolumn{3}{c|}{\textbf{0.25}} & \multicolumn{3}{c|}{\textbf{0.50}} & \multicolumn{3}{c|}{\textbf{0.75}} & \multicolumn{3}{c|}{\textbf{1.25}} & \multicolumn{3}{c|}{\textbf{1.50}} & \multicolumn{3}{c|}{\textbf{1.75}} \\ 
\cmidrule(lr){3-5} \cmidrule(lr){6-8} \cmidrule(lr){9-11} \cmidrule(lr){12-14} \cmidrule(lr){15-17} \cmidrule(lr){18-20}
 &  & ACC & AUR & H-S & ACC & AUR & H-S & ACC & AUR & H-S & ACC & AUR & H-S & ACC & AUR & H-S & ACC & AUR & H-S  \\ 
\midrule
\multirow{11}{*}{\rotatebox[origin=c]{90}{\textbf{CIFAR10-C}}}
& Source  & 81.73 & 77.94 & 79.79 & 81.73 & 77.89 & 79.76 & 81.73 & 77.92 & 79.78 & 81.71 & 77.93 & 79.78 & 81.71 & 77.94 & 79.78 & 81.68 & 78.05 & 79.82 \\
& TENT    & 82.01 & 73.41 & 77.47 & 82.53 & 76.67 & 79.49 & 76.67 & 69.14 & 72.71 & 72.97 & 68.82 & 70.83 & 75.09 & 67.67 & 71.19 & 68.95 & 68.52 & 68.73 \\
& CoTTA   & 87.06 & 83.13 & 85.05 & 86.76 & 85.17 & 85.96 & 86.31 & 85.77 & 86.04 & 85.56 & 86.24 & 85.90 & 85.43 & 86.25 & 85.84 & 85.20 & 86.47 & 85.83 \\
& EATA    & 85.57 & 80.42 & 82.92 & 85.02 & 81.93 & 83.45 & 84.70 & 80.57 & 82.58 & 83.97 & 81.29 & 82.61 & 83.78 & 81.78 & 82.77 & 83.73 & 80.28 & 81.97 \\
& RoTTA   & 83.93 & 79.98 & 81.91 & 84.37 & 81.15 & 82.73 & 84.80 & 82.03 & 83.39 & 84.41 & 81.81 & 83.09 & 83.82 & 81.18 & 82.48 & 83.21 & 80.53 & 81.85 \\
& SoTTA   & 82.56 & 58.15 & 68.24 & 77.72 & 49.72 & 60.64 & 71.93 & 50.23 & 59.15 & 68.35 & 47.53 & 56.07 & 64.67 & 49.80 & 56.27 & 69.52 & 50.21 & 58.31 \\
& OSTTA   & 85.73 & 69.28 & 76.63 & 85.01 & 71.71 & 77.80 & 84.59 & 72.73 & 78.21 & 84.01 & 73.75 & 78.55 & 83.60 & 74.35 & 78.70 & 83.44 & 73.92 & 78.39 \\
& STAMP   & 86.20 & 66.15 & 74.86 & 84.56 & 72.46 & 78.04 & 84.34 & 74.50 & 79.12 & 82.29 & 74.47 & 78.18 & 81.70 & 72.72 & 76.95 & 82.34 & 73.60 & 77.73 \\
& UniEnt  & 85.15 & 90.51 & 87.75 & 85.06 & 91.15 & 88.00 & 84.78 & 90.92 & 87.74 & 84.43 & 89.88 & 87.07 & 84.35 & 89.29 & 86.75 & 84.08 & 88.73 & 86.34 \\
& UniEnt+ & 84.72 & 91.59 & 88.02 & 84.58 & 92.89 & 88.54 & 84.30 & 93.12 & 88.49 & 83.88 & 92.80 & 88.11 & 83.84 & 92.44 & 87.93 & 83.74 & 92.11 & 87.73 \\
\cmidrule(lr){2-20}
& \textbf{Ours} 
& \textbf{88.91}\tiny\color{red}{+1.85} & \textbf{96.45}\tiny\color{red}{+4.86} & \textbf{92.53}\tiny\color{red}{+4.51} 
& \textbf{88.43}\tiny\color{red}{+1.67} & \textbf{97.22}\tiny\color{red}{+4.33} & \textbf{92.61}\tiny\color{red}{+4.07} 
& \textbf{87.90}\tiny\color{red}{+1.59} & \textbf{97.56}\tiny\color{red}{+4.44} & \textbf{92.48}\tiny\color{red}{+3.99} 
& \textbf{87.19}\tiny\color{red}{+1.67} & \textbf{97.32}\tiny\color{red}{+4.52} & \textbf{91.98}\tiny\color{red}{+3.87} 
& \textbf{86.99}\tiny\color{red}{+1.56} & \textbf{96.51}\tiny\color{red}{+4.07} & \textbf{91.50}\tiny\color{red}{+3.57} 
& \textbf{86.48}\tiny\color{red}{+1.28} & \textbf{96.64}\tiny\color{red}{+4.53} & \textbf{91.28}\tiny\color{red}{+3.55} \\
\bottomrule
\multirow{11}{*}{\rotatebox[origin=c]{90}{\textbf{CIFAR100-C}}}
& Source  & 53.25 & 61.02 & 56.87 & 53.25 & 60.76 & 56.76 & 53.25 & 60.63 & 56.70 & 52.85 & 60.48 & 56.41 & 52.87 & 60.49 & 56.42 & 53.00 & 60.55 & 56.52 \\
& TENT    & 41.97 & 64.97 & 51.00 & 31.72 & 69.34 & 43.53 & 29.17 & 67.61 & 40.76 & 22.03 & 66.78 & 33.13 & 22.08 & 67.98 & 33.33 & 21.97 & 68.15 & 33.23 \\
& CoTTA   & 60.07 & 74.25 & 66.41 & 58.88 & 76.39 & 66.50 & 57.79 & 76.25 & 65.75 & 55.72 & 76.49 & 64.47 & 55.59 & 76.58 & 64.42 & 55.58 & 77.21 & 64.63 \\
& EATA    & 61.85 & 71.33 & 66.25 & 59.24 & 73.29 & 65.52 & 55.63 & 72.53 & 62.97 & 52.65 & 69.68 & 59.98 & 52.21 & 70.86 & 60.12 & 50.29 & 69.36 & 58.31 \\
& RoTTA   & 58.62 & 71.45 & 64.40 & 59.14 & 73.32 & 65.47 & 59.32 & 74.61 & 66.09 & 58.77 & 75.19 & 65.97 & 58.42 & 75.13 & 65.73 & 58.36 & 75.20 & 65.72 \\
& SoTTA   & 61.08 & 62.62 & 61.84 & 58.03 & 66.33 & 61.90 & 51.72 & 67.94 & 58.73 & 38.34 & 64.16 & 48.00 & 43.04 & 55.29 & 48.40 & 40.83 & 64.20 & 49.91 \\
& OSTTA   & 62.98 & 74.46 & 68.24 & 61.81 & 75.31 & 67.90 & 60.79 & 75.11 & 67.20 & 58.83 & 75.81 & 66.25 & 58.49 & 75.65 & 65.97 & 58.16 & 75.83 & 65.83 \\
& STAMP   & 64.06 & 54.89 & 59.12 & 61.41 & 47.58 & 53.62 & 61.28 & 90.21 & 72.98 & 58.34 & 91.71 & 71.31 & 57.30 & 91.52 & 70.48 & 56.77 & 90.97 & 69.91 \\
& UniEnt  & 60.89 & 87.56 & 71.83 & 60.43 & 89.47 & 72.14 & 59.96 & 89.31 & 71.75 & 58.53 & 88.02 & 70.31 & 58.14 & 87.80 & 69.96 & 58.04 & 86.96 & 69.62 \\
& UniEnt+ & 60.97 & 88.02 & 72.04 & 60.28 & 90.54 & 72.37 & 59.52 & 91.01 & 71.97 & 57.90 & 90.50 & 70.62 & 57.57 & 90.27 & 70.30 & 57.33 & 89.66 & 69.94 \\
\cmidrule(lr){2-20}
& \textbf{Ours} 
& \textbf{64.83}\tiny\color{red}{+0.77} & \textbf{94.67}\tiny\color{red}{+6.65} & \textbf{76.96}\tiny\color{red}{+4.92} 
& \textbf{64.06}\tiny\color{red}{+2.25} & \textbf{96.74}\tiny\color{red}{+6.20} & \textbf{77.08}\tiny\color{red}{+4.71} 
& \textbf{63.40}\tiny\color{red}{+2.12} & \textbf{97.39}\tiny\color{red}{+6.48} & \textbf{76.80}\tiny\color{red}{+3.82} 
& \textbf{61.59}\tiny\color{red}{+2.66} & \textbf{97.21}\tiny\color{red}{+5.50} & \textbf{75.41}\tiny\color{red}{+4.11} 
& \textbf{61.00}\tiny\color{red}{+2.51} & \textbf{97.21}\tiny\color{red}{+5.69} & \textbf{74.96}\tiny\color{red}{+4.48} 
& \textbf{61.01}\tiny\color{red}{+2.65} & \textbf{96.50}\tiny\color{red}{+5.53} & \textbf{74.76}\tiny\color{red}{+4.82} \\
\bottomrule
\end{tabular}}

\label{tab:sup1}
\end{table}

\end{document}